\documentclass[sigconf,authorversion]{aamas} 

\usepackage{balance} 
\usepackage[utf8]{inputenc}
\usepackage{graphicx}
\theoremstyle{plain}

\theoremstyle{definition}
\newtheorem{defn}{Definition}[section]

\theoremstyle{remark}

\setcopyright{ifaamas}
\acmConference[AAMAS '21]{Proc.\@ of the 20th International Conference on Autonomous Agents and Multiagent Systems (AAMAS 2021)}{May 3--7, 2021}{Online}{U.~Endriss, A.~Now\'{e}, F.~Dignum, A.~Lomuscio (eds.)}
\copyrightyear{2021}
\acmYear{2021}
\acmDOI{}
\acmPrice{}
\acmISBN{}

\acmSubmissionID{45}

\title[Sparse Training Theory for Scalable and Efficient Agents]{Sparse Training Theory for Scalable and Efficient Agents}

\subtitle{Blue Sky Ideas Track}

\author{Decebal Constantin Mocanu}
\affiliation{
  \institution{University of Twente}
  \city{Enschede, the Netherlands}}

\author{Elena Mocanu}
\affiliation{
  \institution{University of Twente}
  \city{Enschede, the Netherlands}}

\author{Tiago Pinto}
\affiliation{
  \institution{Polytechnic Institute of Porto}
  \city{Porto, Portugal}}

\author{Selima Curci}
\affiliation{
  \department{Eindhoven University of Technology}
  \institution{Eindhoven, the Netherlands}}

\author{Phuong H. Nguyen}
\affiliation{
  \department{Eindhoven University of Technology}
  \institution{Eindhoven, the Netherlands}}

\author{Madeleine Gibescu}
\affiliation{
  \department{Utrecht University}
  \institution{Utrecht, the Netherlands}}

\author{Damien Ernst}
\affiliation{
  \department{University of Li{\`e}ge}
  \institution{Li{\`e}ge, Belgium}}

\author{Zita A. Vale}
\affiliation{
  \institution{Polytechnic Institute of Porto}
  \city{Porto, Portugal}}

\begin{abstract}
A fundamental task for artificial intelligence is learning. Deep Neural Networks have proven to cope perfectly with all learning paradigms, i.e. supervised, unsupervised, and reinforcement learning. Nevertheless, traditional deep learning approaches make use of cloud computing facilities and do not scale well to autonomous agents with low computational resources. Even in the cloud, they suffer from computational and memory limitations, and they cannot be used to model adequately large physical worlds for agents which assume networks with billions of neurons. These issues are addressed in the last few years by the emerging topic of sparse training, which trains sparse networks from scratch. This paper discusses sparse training state-of-the-art, its challenges and limitations while introducing a couple of new theoretical research directions which has the potential of alleviating sparse training limitations to push deep learning scalability well beyond its current boundaries. Nevertheless, the theoretical advancements impact in complex multi-agents settings is discussed from a real-world perspective, using the smart grid case study.
\end{abstract}


\newcommand{\BibTeX}{\rm B\kern-.05em{\sc i\kern-.025em b}\kern-.08em\TeX}

\begin{document}


\pagestyle{fancy}
\fancyhead{}


\maketitle 


\section{Introduction}

In recent years, deep learning has become an alternative name for artificial intelligence (AI), while many other fields have remained behind. One one side, this is due to the success of deep learning in solving some real-world open problems in computer vision and natural language processing~\cite{lecun2015deeplearning, NIPS2017_3f5ee243}, and on the other side due to the slower pace of advancement made by the other AI fields, including here the world of autonomous agents~\cite{agents_dead_live_aamas2019}. Nevertheless, one of the main goals of autonomous agents is to behave \textit{intelligently} and to be able to learn and reason in order to take optimal decisions and actions based on their perception of their world (environment). 

The main difficulty in using deep learning capabilities in the field of autonomous agents is that deep learning needs extensive computing facilities in the cloud and almost production-ready software libraries to work properly. The high computational cost of deep learning \cite{jouppi2017datacenter}, given mainly by the very high number of parameters (or connections) in dense neural networks (NNs), prohibits the agents which many times run in low-resource (or embedded) devices to properly make use of the deep learning architectures -- including here the capacity of (re)training the model directly in the agent environment without the need of the cloud infrastructure. Even the agents running in the cloud suffer and are limited by the high deep learning costs. Up to our best knowledge, these problems have started to be preliminary discussed in our AAMAS 2019 tutorial~\cite{AAMAS2019tutorial}. Herein, we take these ideas to the next level, by starting to formalize the theory and by formulating the open research questions, challenges, and possible solutions.

Firstly, this paper advocates that alternative solutions for the high computational costs of traditional deep learning paradigms have started to emerge, i.e. the sparse-to-sparse training paradigm (or, on short, sparse training)~\cite{anonymous2021selfish} to train sparse NNs from scratch \cite{Mocanu2016xbm,mocanu2018scalable,dettmers2019sparse}. Auxiliary, and not fully understood yet, this paradigm may also bring better performance and generalization power than dense NNs as empirically observed by~\cite{ICML-2019-PetersonB0GR,evci2020gradient,liu2020topological}. Secondly, the paper presents the advantages of using sparse training for autonomous agents and discusses the main challenges which have to be solved in order to take full benefits of these advantages. To the end, the paper analyzes hypothetically intelligent agents enhanced with sparse training capabilities in real-world complex systems. 
\section{Sparse Neural Networks}
\begin{figure}[ht!]
    \centering
    \includegraphics[width=1\linewidth]{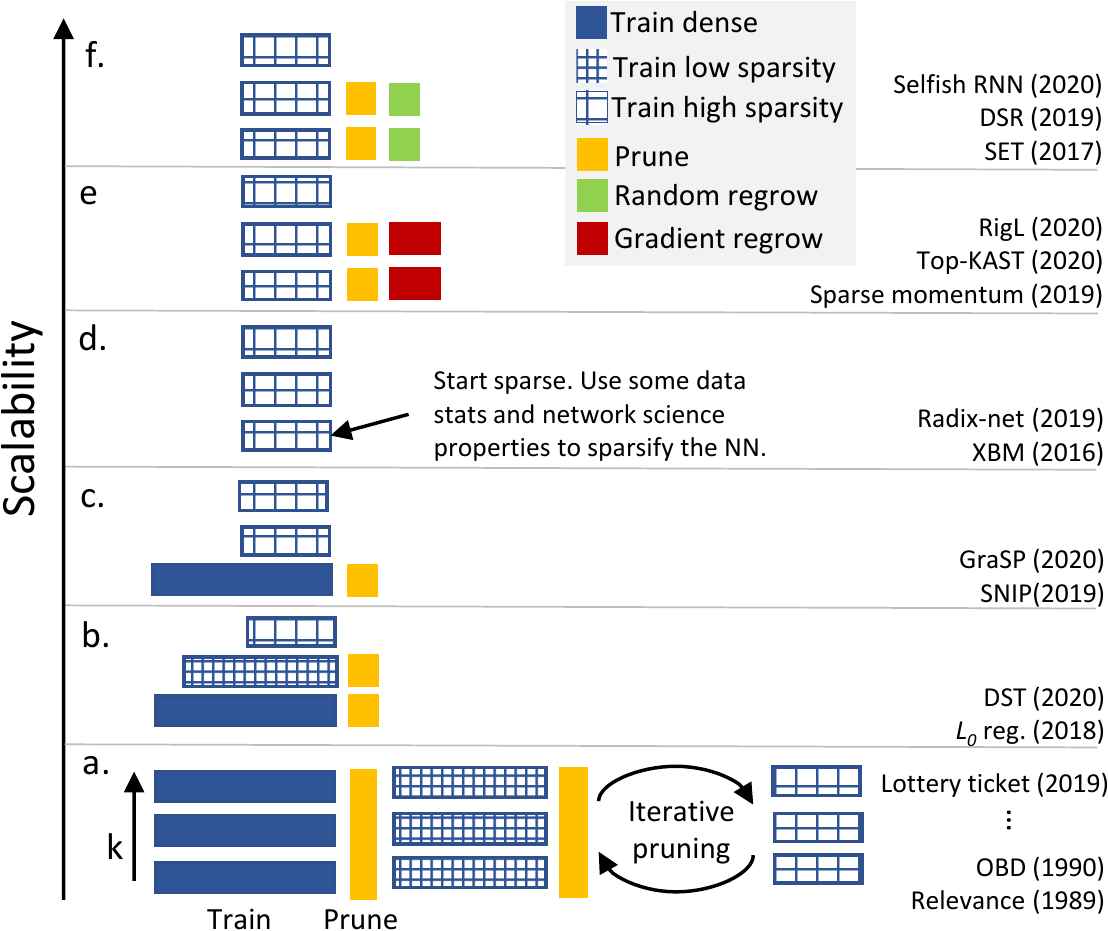}
    \caption{Schematic representation of various method types used to obtain sparse neural networks and a rough estimation of their scalability; a. Pruning, b. Simultaneously training and pruning, c. One-shot pruning, d. Sparse training (static), e. Sparse training (dynamic - gradient), f. Sparse training (dynamic - random).}
    \label{fig:sparsetraining}
\end{figure}
\subsection {Definitions}
A typical neural network relies on a dataset $\mathcal{D}$ of labeled pairs $(x, y)$ drawn from a distribution $D$ over the input/output spaces $X$ and $Y$. The goal is to use $\mathcal{D}$ to search a neural network function parameterized by $\phi$, with $\phi_w \in \mathbb{R}^d$ that has low expected test loss $l(\phi_w (x), y)$ when using $x$ to predict the associated $y$ on unseen samples drawn from $\mathcal{D}$, as measured by the loss $l : Y \times Y \to [0, \infty)$.

\begin{defn}
A dense neural network has an underlying connectivity graph $G(\mathcal{V}, \mathcal{E})$, with $\mathcal{V}$ representing the set of vertices (nodes or neurons) and $\mathcal{E}$ representing the set of edges (connections or parameters) composed by many densely -- even fully -- connected components organized in layers. 
\end{defn}
\begin{defn}
A sparse neural network has an underlying connectivity graph $G(\mathcal{V}, \mathcal{E'})$ induced on the same set $\mathcal{V}$ of nodes and a subset $\mathcal{E}'\subset \mathcal{E}$ of edges, where the cardinal number of set $|\mathcal{E'}|\ll|\mathcal{E}|$, leading to a much sparser components connectivity.
\end{defn}

There is a myriad of options to define the graph G underlying a NN and this makes it very difficult to study a general form of it. For simplicity, and without the loss of generality, further we will refer to the graph $G(\mathcal{V}, \mathcal{E})$ as one of the most used and simple component of a NN, i.e. a fully-connected (FC) bipartite graph between two consecutive layers of neurons.

\subsection{Dense training}
 In the case of FC bipartite layers, $|\mathcal{E}|\approx|\mathcal{N}|^2$. In practice, training a dense NN on $\mathcal{T} \in \mathcal{D}$ data-points and a batch size $b$ requires a large number of weights updates. A typical NN training procedure using Stochastic Gradient Descent would perform a series of weight updates for each parameter $\phi$ of the network (the elements of $\mathcal{E}$) to minimize the loss $l$. Let's note $k \in \mathbb{N}$ the hyperparameter reflecting how many times the training set is parsed (in other words, the number of training epochs) until $l$ converges. The number of weight updates until convergence is given by:
 \begin{equation}
\textstyle k~\frac{|\mathcal{T}|}{b}|\mathcal{E}|  
\end{equation}

\subsection{Dense-to-sparse training}
The first class of methods used to obtain sparse neural networks offers benefits just in the inference phase. Broadly speaking, they start training from densely connected networks followed by pruning of unimportant connections.

\textbf{Traditional iterative pruning (Fig. \ref{fig:sparsetraining}a).} 
Aiming to reduce the computational complexity of a NN, a large amount of research was focused on~\textit{pruning}. The term was coined in earlier '90 by Mozer and Smolensky~\cite{MozerSmolensky1989pruning} and LeCun et al.~\cite{lecun1990optimal} who proposed Optimal Brain Damage (OBD). It introduces the idea of first training completely a dense neural networks, then prune unimportant connections, and after that train and prune successively for $p$ steps the remaining sparse network until a trade-off between  performance and network size is reached. Han et al.~\cite{han2015learning} refined the idea in the context of deep neural networks, while the Lottery Ticket Hypothesis~\cite{frankle2018the} shows that it is sufficient to set $p=2$ if the pruned networks is trained from random initialized weight values. The typical number of weight updates is given by:
 \begin{equation}
\textstyle\sum_p k~\frac{|\mathcal{T}|}{b}|\mathcal{E}_p^{'}|, \text{where }\mathcal{E}_0^{'}=\mathcal{E} \text{ such that }\mathcal{E}_{p+1}^{'}\subset \mathcal{E}_p^{'}, \forall p
\end{equation}

\textbf{Simultaneously training and pruning (Fig. \ref{fig:sparsetraining}b).} Louizos et al.~\cite{louizos2017learning} proposed in 2018 a slightly more efficient procedure which starts from a dense NN and trains and prunes simultaneously the network parameters using $L_0$ regularization. DST revised the idea in \cite{LIU2020Dynamic}. Typically, the total number of weight updates are given by:

 \begin{equation}
\textstyle\sum_k \frac{|\mathcal{T}|}{b}|\mathcal{E}_k^{'}|, \text{where }\mathcal{E}_0^{'}=\mathcal{E} \text{ such that }|\mathcal{E}_{k+1}^{'}|<|\mathcal{E}_k^{'}|, \forall k
\end{equation}

\textbf{One-shot-pruning (Fig. \ref{fig:sparsetraining}c).} 
SNIP in 2019 ~\citep{lee2018snip} and GrasSP in 2020 \cite{Wang2020GraSP} have shown that it is possible to prune many large NNs faster, by training the dense network just for a very short period of time (not until convergence) on a data subset $\mathcal{T}' \subset \mathcal{T}$. To obtain also a decent performance the training process is continued on the remained sparse network on the whole dataset $\mathcal{T}$ \cite{Zhang2020One-Shot}. Approximately, the number of weight updates is:
 \begin{equation}
\textstyle\sum_i\frac{|\mathcal{T'}|}{b}|\mathcal{E}| + \sum_k \frac{|\mathcal{T}|}{b}|\mathcal{E}^{'}|,  \text{where }\mathcal{T}^{'}\subset\mathcal{T} \text{, }  i\ll k \text{, }  |\mathcal{E}{'}|\ll|\mathcal{E}| 
\end{equation}

\subsection{Sparse-to-sparse training}
In the last years, many works have started to study also the efficiency of the training phase by training end-to-end sparse NNs\footnote{It worth mentioning that the term "sparse neural networks" appears in arXiv pre-prints title and abstracts as follows: between 2010 and 2017 just one or two papers per year contain it; 2018 - 9 papers; 2019 - 11 papers; and 2020 - 27 papers. Of course, other terms may refer to the same concept, and by relaxing the search query, we obtain 128 papers on the topic in 2020, in comparison with 76 in 2019.}. The concept recently has started to be known as \textit{sparse training}. 

\textbf{Static sparse connectivity (Fig. \ref{fig:sparsetraining}d)} 
In 2016, Mocanu et al. \citep{Mocanu2016xbm} have introduced the idea of training directly from scratch (without pruning from a dense networks) sparse Restricted Boltzmann Machines (named XBMs) by using static sparsity. The concept has been further studied in \citep{kepner2019radix} (i.e. Radix-net) and ~\citep{prabhu2018deep, ailon2020sparse, anonymous2021keep} for other NN types. The connectivity graph is obtained before training with data statistics, network science and graph theory concepts. After that, the sparse NN is trained normally using gradient descent or an alternative method. The number of weight updates can be approximated with:
\begin{equation}
\textstyle k~\frac{\mathcal{T}}{b}|\mathcal{E}^{'}|  \quad \text{such that } |\mathcal{E}{'}|\ll|\mathcal{E}| 
\end{equation} 
The main disadvantage of sparse NNs with static sparsity is that the connectivity pattern is designed by hand, and cannot model very well data distribution. This leads, in general, to a relatively weaker performance than their dense counterparts.

\textbf{Dynamic (adaptive) sparse connectivity.} \newline
\textit{Random regrow (Fig. \ref{fig:sparsetraining}f).} To alleviate this problem, Mocanu et al. \cite{decphdthesis, mocanu2018scalable} have introduced in 2017 and 2018, respectively, the key concept of adaptive (or dynamic) sparsity and the Sparse Evolutionary Training (SET) algorithm in order to automatically discover an optimal sparse connectivity pattern during training. On very short, SET starts from a randomly generated sparse connectivity and uses after each training epoch a prune-and-regrow strategy to automatically model the data distribution and to find an optimal sparse connectivity. More exactly, during each prune-and-regrow step, a percentage of the non-important connections (the ones closest to zero) are pruned, and the same amount of pruned connections is added to the resulted sparse network in randomly selected positions. SET reduces quadratically the number of connections, while obtaining better accuracy than the dense counterparts in the case of MultiLayer Perceptron (MLP). Further on, \cite{bellec2018deep} trained very sparse networks by sampling the sparse connectivity based on a Bayesian posterior, while ~\cite{mostafa2019parameter} introduced Dynamic Sparse Reparameterization (DSR) to train Convolutional Neural Networks which dynamically adjust the sparsity levels of different layers. Based on SET and DSR, \cite{anonymous2021selfish} proposed a specialized sparse training method for sparse Recurrent Neural Networks (RNNs), named Selfish RNN, which outperforms its dense counterpart at reasonable sparsity levels
(less sparser than in the case of MLPs).

\textit{Gradient regrow (Fig. \ref{fig:sparsetraining}e).} To enhance a faster convergence, \cite{dettmers2019sparse} and \cite{evci2019rigging} introduced the idea
of using momentum and gradient information (quantified in two methods named, Sparse Momentum and RigL, respectively) from non-existing connections during the regrow steps. Evci et al. \cite{evci2019rigging} show that if they trained with RigL sparse CNNs for a long enough time, they can reach the performance of the dense counterparts. Nevertheless, using information from non-existing connections is not scalable, and~\cite{Jayakumar2020TopK} improves RigL by considering just the gradients of a subset of the non-existing connections in Top-KAST.  For both, random and gradient regrow, the number of weight updates can be approximated with:
\begin{equation}
\textstyle k~\frac{\mathcal{T}}{b}|\mathcal{E}_k^{'}| \quad \text{such that } |\mathcal{E}_{k}^{'}|=|\mathcal{E}_{k-1}^{'}|\propto|\mathcal{N}|\text{ and }  \mathcal{E}_{k-1}^{'}\neq\mathcal{E}_{k}^{'}   
\end{equation} 

\section{Sparse Training Agents}
While sparse training seems a promising research direction (as reflected also by Equations 1-6), there are many open questions waiting to be answered and challenges which have to be solved in order to obtain real scalable NN solutions to be further used in any form of intelligent agent. We group them in two main categories: hardware and software support, and theoretical open questions.
\subsection{Hardware and software support}
Due to the relatively small number of parameters in comparison with the dense models, theoretically, sparse training allows a jump with few orders of magnitude in the representational power of NNs, while having better generalization capabilities than the dense counterparts~\cite{ICML-2019-PetersonB0GR,anonymous2021gradient,liu2020topological}. These have the potential of making sparse training the \textit{de facto} approach for NNs in the future. The main obstacle in making this jump is that most deep learning specialized hardware is optimized for dense matrix operations, and it practically ignores sparse matrix operations. Due to this reason, almost all research done in sparse NNs uses a binary mask over weights to simulate sparsity. In terms of hardware,~\cite{7783723, 9025249} have started paving the ground for sparsity proper hardware support. Also, it worth mentioning NVIDIA A100, released in May 2020, which supports 2:4 sparsity (i.e. 50\% fixed sparsity level)~\cite{anonymous2021learning}. These suggest that more hardware support will be developed for sparse NNs in the medium-term future.
While waiting for new hardware with proper support for sparse NNs to be conceived, e.g.~\cite{predifinedsparsehardware2019}, for both edge and cloud, the problem can be addressed as a programming and software engineering challenge. Indeed, mainstream libraries are not optimized for sparse matrix multiplication, but taking an alternative approach and looking to the basic programming and data structure principles, we can devise new parallel algorithms and implementations for CPUs and GPUs to enable truly sparse NNs. Up to our knowledge, there are very few works on this topic. 

For training efficiency, Liu et al.~\cite{liu2020sparse} have been able to build truly sparse MLPs with over one million neurons on a typical laptop using only basic Python libraries. In~\cite{atashgahi2020quick}, a truly sparse denoising autoencoder is proposed in order to perform fast and robust feature selection. Interestingly, when the number of neurons is sufficiently large, this sparse autoencoder running just on one CPU core has a faster training time than its dense counterparts running on GPU, while being able to perform better feature selection and to have a much more environmentally friendly energy consumption.
For inference efficiency, it worth mentioning here the recent MIT/IEEE/Amazon Sparse Deep Neural Network Challenge \cite{Kepner_2019}, for which the authors of \cite{sparseinference} have been able to perform inference (no training at all) with 125 million neurons in random networks.
Nevertheless, all of the above can be reduced to two practical unanswered questions:
\begin{itemize}
    \item How much we can improve the \textit{training and inference execution time} using sparse NNs instead of dense NNs? 
    \item Where is the trade-off between speed and performance in the case of sparse neural networks?
\end{itemize}

\subsection{Theoretical open questions}
 \subsubsection{Understanding Sparse Training at Scale} Up to now, no one had the opportunity to study any form of NN with billions of neurons (to model, for instance, large realistic worlds), partly also because the algorithmic novelty is too much driven nowadays by the hardware limitations~\cite{hooker2020hardware} and the ease of using almost production-ready deep learning solutions. In \cite{liu2020topological}, it has been demonstrated that a vast number of entirely different sparse topologies, unveiled easily using dynamic sparsity, reach a very similar performance level after training. Nevertheless, the effect of sparsity on various NN models is non-proportional and non-derivable directly from their dense counterparts. It seems that there is no simple unified way of treating sparsity similarly in MLPs, RNNs, CNNs, and so on. Altogether, this leads to several questions:
 \begin{itemize}
     \item To what extend sparse training applied to sparse NNs with billions of neurons will resemble its learning behaviour observed currently in networks with thousands of neurons?
     \item How can we combine multiple sparse NNs into one for parallel sparse training algorithms or federated learning \cite{sparsefedlearning2020}? 
    \item Dynamic sparsity naturally can help with regularization, but what makes it to lead to better generalization as empirically shown, for instance, in~\cite{ICML-2019-PetersonB0GR}? How is this even possible as, by design, sparse NNs do not benefit of overparameterization? \item To what extend the capacity of sparse training to outperform dense NNs is sensitive to the NN model type and settings?
        \item Is dynamic sparsity in sparse training always converging if we consider that it addresses a combinatorial and a continuous optimization problem simultaneously? 
    \item Can we formulate in a trustable and understandable format the relations between the
    output and the input data in a sparse neural networks using its sparse graph properties? 
\end{itemize}  
    
\subsubsection{Sparse Training and Reinforcement Learning} 
At the core of agents' world is reinforcement learning. Up to now, sparse training has been developed and studied just in the context of supervised and unsupervised learning. To enhance Deep Reinforcement Learning (DRL) algorithms with sparse training, some fundamental challenges have to be solved:
\begin{itemize}
\item Can we create sparse DRL models, where the NNs used as function approximators are NNs with dynamic sparsity?
\item To what extend the above-created models will be successful as their adaptive connectivity would have to cope also with the dynamicity and uncertainty of the DRL environments?
\item Are there any theoretical guarantees regarding the convergence of such sparse DRL models?
\end{itemize}

\subsubsection{Realistic Continual Learning at Scale} Currently, continual learning, even if it is considered a very important topic by the research community, it is immature. For instance, we do not have good enough algorithms to outperform in continual learning settings single task learners. With increased representational power, sparse networks represent a natural candidate in implementing and improving continual learning \cite{sokar2020spacenet, wortsman2020supermasks}. 
In order to have autonomous intelligent agents continuously learning, next question arises:
\begin{itemize}
 \item To what extend sparse training can help in learning a long series of tasks without suffering from catastrophic forgetting and without storing all historical data?
  \end {itemize}

\section{Real-world complex systems}
\textbf{Neural Networks applications at Scale.} Billions of neurons will open many new possibilities of exploiting NNs, well beyond what we know and imagine today. The natural question arising is: \textit{What can we do with them?} For instance, ~\cite{liu2020sparse} showed the benefit of sparse training for microarray data and its potential of replacing the traditional two steps approach (feature selection, followed by classification). Thus, we would have to rethink the utility of such large networks ahead of time. 

\textbf{Multi-agent systems.} Imagine, for instance, myriads of autonomous and intelligent agents perceiving, cooperating, and acting in a realistic environment to fulfill some common goals (e.g. safe and reliable autonomous driving). Many of these heterogeneous agents would have by design and hardware constraints limited capabilities, but in order to properly model and interact among them and with the environment some of them would need to have extremely large representational and generalization capacity.

\textbf{Autonomous NNs in Low-Resource Devices.} At the other end of the spectrum, by using sparsity, we can devise NNs which can be entirely independent of the cloud and can be trained directly on low-resource devices. Such a scenario can be useful for many reasons, for instance, data privacy, energy-saving, harsh, remote environments (e.g. IoT sensors in the ocean), communication overheads or failures in autonomous driving, and so on. 
 \begin{figure}
\centering
\includegraphics[width=0.45\textwidth]{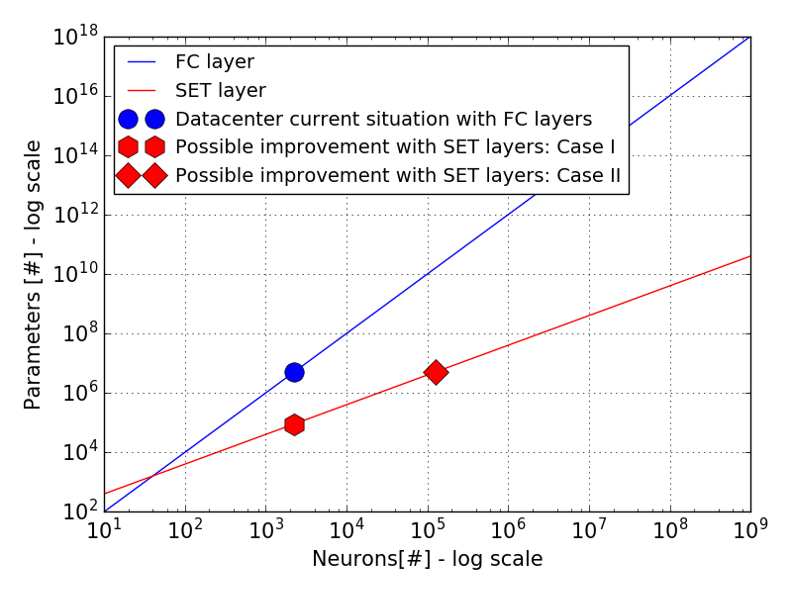}
\caption{Sparse NN hypothetical improvement for \cite{jouppi2017datacenter}.} 
\label{fig:datacenter}
\end{figure}

 \textbf{The Smart Grid case study.} Large and complex systems, such as electricity grid, relies on the multi-agents learning capabilities to act in an accurate, fast, cheap, reliable and trustworthy manner helping the transition towards a more human-centred system at the local level. Still, the multi-agents coordination is needed at all levels~\cite{KoenAAMAS2010}. Specialized agents, using DRL have been proved to have superior learning capabilities. For example, the building resource allocation problem typically requests two steps: a) building energy prediction, and b) building energy optimization. In~\cite{Mocanu2019DRL} it has been shown that it is possible to use DRL to replace both, the prediction an optimization steps, with an on-line adaptive solution. In~\cite{Zhand2020overviewDRLenergy}, a detailed overview of the DRL application in the energy system highlights the huge interest, as well as a general recommendation to overcome the scalability open problems. 
Even if in 2020, an edge computing hardware prototype \cite{GebbranAAMAS2020EdgeComputingHardwarePrototype} for coordination of prosumer agents appeared, the most advanced multi-agent based real-time energy infrastructure~\cite{Tiago2020AAMAS} and specialized agents~\cite{Habibi2015stateestimation, Mocanu2019DRL} are relying on cloud computing and do not scale~\cite{Weerdt2018NotScale,Zhand2020overviewDRLenergy}. 
We argue in the favour of a further investigation of sparse DRL in multi-agent systems with respect to different types of strategies as a way to infuse more scalable AI technology into the energy field and help the ongoing transition towards a more sustainable society.

 On the other side, it is good to highlight the tremendous energy costs induced by the deep learning approaches. For example, Fig. \ref{fig:datacenter} presents a typical dense MLP model from the Google Datacenter (representing 61\% of the load) \cite{jouppi2017datacenter} and hypothesizes how a sparse MLP may improve the performance in two cases: (I) much faster training and inference time, lower memory, cost reduction, and probably better accuracy; (II) much higher representational power.
 
At the same time, we shall not forget that if we want to solve problems such as climate change~\cite{Vahid2020climate}, we shall not look to AI just as to a problem solver, but also to try avoiding making AI a burden of the energy system. Thus, cost-efficient and environmental friendly AI models shall be an improvement goal in itself,  considering \textit{sparse AI for energy and energy for sparse AI}.

\section{Discussion}

All the directions and questions presented in the paper may have a bit of a spaghetti effect at a quick look. Although, probably this is the only solution to obtain intelligent agents and heterogeneous systems of agents which can interact among them in order to solve vital societal challenges and to contribute to the advancement of artificial intelligence. We believe that a way to move forward is to study these challenges as a whole within the research community. Consequently, instead of a concluding phrase, by listing an arguable quote from Metaphysics of Aristotle: \textit{The whole is more than the sum of its parts}, we invite researchers to brainstorm with us.

\bibliographystyle{ACM-Reference-Format} 
\bibliography{sample}


\end{document}